\documentclass[conference]{IEEEtran}
\IEEEoverridecommandlockouts 
\usepackage[utf8]{inputenc} 
\usepackage{amsmath,amssymb,amsfonts}
\usepackage{graphicx}
\usepackage{cite}
\usepackage{hyperref}
\usepackage[nameinlink]{cleveref}
\crefname{figure}{Fig.}{Figs.}
\Crefname{figure}{Fig.}{Figs.}
\usepackage{booktabs}
\usepackage{subfig}

\begin{document}

\title{Dynamic Lookahead Distance via Reinforcement Learning-Based Pure Pursuit for Autonomous Racing}

\author{
Mohamed Elgouhary and Amr S. El-Wakeel
\thanks{The Authors are with the Lane Department of Computer Science and Electrical Engineering, West Virginia University, Morgantown, WV, USA.
        {\tt\small mae00018@mix.wvu.edu, amr.elwakeel@mail.wvu.edu}}%
\thanks{This work was partially supported by DARPA AI-CRAFT under Grant AWD16069.}%
}

\maketitle

\begin{abstract}
Pure Pursuit (PP) is a widely used path-tracking algorithm in autonomous vehicles due to its simplicity and real-time performance. However, its effectiveness is sensitive to the choice of lookahead distance: shorter values improve cornering but can cause instability on straights, while longer values improve smoothness but reduce accuracy in curves. We propose a hybrid control framework that integrates Proximal Policy Optimization (PPO) with the classical Pure Pursuit controller to adjust the lookahead distance dynamically during racing. The PPO agent maps vehicle speed and multi-horizon curvature features to an online lookahead command. It is trained using Stable-Baselines3 in the F1TENTH Gym simulator with a KL penalty and learning-rate decay for stability, then deployed in a ROS2 environment to guide the controller. Experiments in simulation compare the proposed method against both fixed-lookahead Pure Pursuit and an adaptive Pure Pursuit baseline. Additional real-car experiments compare the learned controller against a fixed-lookahead Pure Pursuit controller. Results show that the learned policy improves lap-time performance and repeated lap completion on unseen tracks, while also transferring zero-shot to hardware. The learned controller adapts the lookahead by increasing it on straights and reducing it in curves, demonstrating effectiveness in augmenting a classical controller by online adaptation of a single interpretable parameter. On unseen tracks, the proposed method achieved 33.16\,s on Montreal and 46.05\,s on Yas Marina, while tolerating more aggressive speed-profile scaling than the baselines and achieving the best lap times among the tested settings. Initial real-car experiments further support sim-to-real transfer on a 1:10-scale autonomous racing platform.
\end{abstract}

\begin{IEEEkeywords}
Autonomous Racing, Reinforcement Learning, Pure Pursuit, PPO, Dynamic Lookahead, Roboracer, Stable-Baselines3
\end{IEEEkeywords}

\section{Introduction}
Autonomous racing is a rapidly evolving domain that tests the limits of perception, planning, and control algorithms under high-speed and high-precision requirements. Unlike traditional autonomous driving, racing involves aggressive maneuvers, tight curves, and the need for real-time adaptation to dynamic vehicle and track conditions. Designing controllers that can operate safely and efficiently in such environments remains a central challenge in robotics and intelligent systems.

Among classical trajectory tracking strategies, Pure Pursuit \cite{Pure-Pursuit} has emerged as a widely adopted geometric controller due to its conceptual simplicity, low computational overhead, and reasonable performance in many driving scenarios. It works by projecting a lookahead point on a reference trajectory and computing a steering command that aims the vehicle toward that point. However, despite its popularity, the Pure Pursuit controller exhibits a fundamental limitation: its behavior is highly sensitive to the choice of the lookahead distance.

The lookahead distance determines the trade-off between stability and responsiveness. A small lookahead value can improve curve tracking and allow faster directional changes, but it often causes steering oscillations and instability on straight segments. A large lookahead value usually produces smoother and more stable motion, but it can cut corners and perform poorly in sharp turns. In fixed-lookahead implementations, the controller cannot adapt to changing track conditions, which leads to suboptimal performance, especially in high-speed racing where both accurate cornering and stable straight-line behavior are important.

Several heuristic approaches have attempted to address this issue, such as curvature-aware or speed-dependent lookahead modulation. These methods can improve performance in specific scenarios, but they still rely on manually crafted rules and schedule parameters that often require retuning across tracks, speed profiles, and vehicle platforms. In contrast, a learned adaptation policy can potentially capture more context-dependent behavior while retaining the structure of the underlying controller.

To overcome these limitations, we propose a hybrid control framework that augments classical Pure Pursuit with a learning-based adaptation mechanism \cite{Learning-Based}. Specifically, we use Proximal Policy Optimization (PPO) \cite{PPO} to dynamically adjust the lookahead distance in real time based on the vehicle's current state. Unlike fixed tuning and hand-crafted adaptive schedules, our approach formulates lookahead selection as a continuous control task, enabling the policy to learn context-dependent behaviors while preserving the structure and interpretability of the underlying geometric controller.

The PPO agent is trained in simulation using the F1TENTH Gym environment \cite{f1tenth}, where it learns to map vehicle speed and multi-horizon raceline curvature features to an optimal lookahead distance. The integration of PPO with the Pure Pursuit controller ensures that learning remains focused on adaptive behavior while retaining the interpretability, robustness, and low-latency characteristics of classical control. We do not replace the underlying geometric controller or learn low-level steering and throttle end-to-end. Instead, we focus on a narrower and more interpretable problem: adapting the Pure Pursuit lookahead distance online while preserving the structure, transparency, and low computational cost of the classical controller. This framing is especially useful in modular ROS2-based autonomy stacks, where bounded computation and controller interpretability remain important.

To support the learning process, we leveraged a globally optimized race trajectory to extract reference waypoints \cite{tum}, including $x$, $y$ positions, target speed, and curvature at each point. These were used both to guide the baseline controller and to define relevant state features for the PPO agent. The agent observes vehicle speed and multi-horizon curvature features to learn a dynamic mapping to the lookahead distance. LiDAR is used for collision detection in the environment and reward to enforce safety during training.

For efficient and stable training, we utilized Optuna \cite{Optuna} to optimize key hyperparameters, including the initial learning rate and maximum gradient norm. We further stabilized training by applying KL-divergence penalties and learning rate decay, enabling smoother convergence and improved final performance.


Because the learned policy adjusts only the Pure Pursuit lookahead parameter while leaving the geometric controller unchanged, the resulting method remains lightweight, interpretable, and easy to integrate into modular ROS2 autonomy stacks.


Our contributions are as follows:
\begin{itemize}
    \item We introduce a reinforcement learning-augmented Pure Pursuit controller that dynamically adjusts the lookahead distance online for autonomous racing.
    \item We design a PPO training setup based on speed and multi-horizon curvature features, together with a reward that promotes progress, smoothness, and safe lap completion.
    \item We evaluate the learned policy zero-shot on unseen tracks against both fixed-lookahead and hand-crafted adaptive Pure Pursuit baselines, and further validate sim-to-real transfer on a real 1:10-scale autonomous racing vehicle against a fixed-lookahead Pure Pursuit baseline.
    \item We adapt a single interpretable Pure Pursuit parameter ($L_d$), yielding behavior consistent with racing intuition while preserving the classical controller structure.
\end{itemize}

This work highlights the potential of combining the adaptability of deep reinforcement learning with the reliability and transparency of classical control. The resulting controller is well-suited for high-performance autonomous racing and represents a step toward more intelligent, context-aware control strategies in robotics.

\section{Related Work}
Path tracking is a foundational problem in autonomous driving, with numerous classical \cite{path-tracking} and learning-based approaches \cite{Learning-Based} developed over the years. Among classical methods, geometric controllers such as Pure Pursuit \cite{Pure-Pursuit}, Stanley \cite{Stanley}, and the Follow-the-Carrot algorithm have gained widespread popularity due to their simplicity and ease of implementation. In particular, the Pure Pursuit algorithm has seen extensive use in autonomous ground vehicles for applications ranging from urban navigation \cite{urban}, \cite{urban2} to racing \cite{racing}, due to its ability to operate in real time with minimal computational resources.

Several enhancements to the Pure Pursuit controller have been proposed to improve its adaptability to varying driving conditions. One common direction involves dynamically adjusting the lookahead distance based on vehicle speed, track curvature \cite{curve}, \cite{curve2}, or a combination of both. For example, curvature-aware lookahead modulation aims to reduce the lookahead in tight curves to improve responsiveness and increase it on straights for stability. Speed-based modulation similarly attempts to increase the lookahead at higher velocities to prevent overreacting to small deviations. While effective in specific settings, these heuristics are typically hand-tuned and may not generalize across different vehicle models, tracks, or driving scenarios.

On the other hand, reinforcement learning (RL) has emerged as a powerful tool for developing adaptive and high-performance control policies \cite{RL}. Deep RL algorithms such as Deep Q-Networks (DQN) \cite{DQN}, Deep Deterministic Policy Gradient (DDPG) \cite{DDPG}, and Proximal Policy Optimization (PPO) \cite{PPO} have been applied to a wide range of robotic control problems. Within the domain of autonomous driving, end-to-end learning approaches have demonstrated promising results \cite{End-to-End}, where policies learn directly from sensory input to output control commands. However, end-to-end policies that output low-level actions can be harder to analyze and validate for deployment, and may degrade under distribution shift (e.g., new tracks or sim-to-real mismatch) \cite{challenges,safety,sim}.

In the context of autonomous racing, recent work has explored the use of RL for tasks such as trajectory planning, throttle control, and full vehicle control under competitive conditions. Environments such as Roboracer \cite{f1tenth} have become common testbeds for these methods. In particular, learning-based approaches have demonstrated their ability to outperform classical controllers in aggressive racing scenarios \cite{comparison}. However, most of these methods replace classical control entirely, often at the cost of interpretability and robustness.

Our work lies at the intersection of these two paradigms. Rather than replacing the classical controller, we use reinforcement learning to augment it by tuning a critical hyperparameter---the lookahead distance---based on the current driving context. This hybrid approach combines the adaptability of RL with the proven stability and interpretability of geometric control \cite{combination}. Similar hybrid frameworks have been explored in related domains, such as adaptive PID tuning \cite{pid} and gain scheduling via RL \cite{gain}. However, few have applied this concept to geometric path tracking in autonomous racing \cite{Pure-Pursuit}.

In this study, we compare against both fixed-lookahead Pure Pursuit and a hand-crafted adaptive Pure Pursuit baseline that adjusts lookahead online using rule-based scheduling. This allows us to distinguish the benefit of learning-based adaptation from the benefit of dynamic lookahead alone. While related work has explored reinforcement learning for dynamic lookahead generation in truck-driving environments \cite{truck}, our formulation targets the distinct demands of high-speed autonomous racing. The proposed approach retains the strengths of model-based control while enabling data-driven adaptation, making it well-suited for real-time, high-speed, and safety-critical scenarios. In addition to local path-tracking strategies, global trajectory optimization has been employed in autonomous racing to precompute an optimal racing line for the entire track. The minimum-curvature method, as implemented in the TUM \texttt{global\_racetrajectory\_optimization} framework \cite{tum}, generates a smooth path within track boundaries by minimizing integrated squared curvature. This approach produces racing lines that balance stability and cornering capability, and is often used as a reference trajectory for downstream controllers such as MPC or Pure Pursuit.

\section{Background and Proposed Method}
\subsection{Background Components}
\noindent This section summarizes our end-to-end pipeline. We localize with a LiDAR-based MCL particle filter, generate a minimum-curvature racing line, and overlay a PPO-trained policy that selects a \emph{dynamic} Pure Pursuit lookahead online. We then detail the observation/action spaces, reward, models, training setup, and how these components integrate in simulation and on the real RoboRacer car.

\subsection{Monte Carlo Localization via Particle Filter}
We estimate global pose with a Monte Carlo Localization (MCL) particle filter \cite{mcl,pf} that uses LiDAR measurements and a precomputed occupancy grid. The filter maintains a set of weighted pose hypotheses (“particles”) that approximates the posterior over the vehicle’s $(x,y,\psi)$.

At each time step, every particle is propagated forward using a bicycle-model kinematics driven by the commanded speed and steering. Small, zero-mean process noise is injected to capture unmodeled dynamics and odometry uncertainty.

For each particle, we ray-cast the occupancy grid to predict LiDAR ranges and compare them with the observed scan using a standard beam-based likelihood model. Particles that better explain the scan receive higher weights.

When weights become overly concentrated (low effective sample size), we apply systematic resampling to refocus particles around high-probability regions while limiting particle impoverishment.

The vehicle pose is read out as the weighted average of the particle set; the heading is computed via a circular (angle-aware) mean. This implementation follows the classical formulation of Thrun, Burgard, and Fox and the MIT notes.

\subsection{Global Trajectory Generation via Minimum Curvature Optimization}
To generate the reference trajectory for our controller, we adopt the \emph{minimum curvature} method implemented in the TUM \texttt{global\_racetrajectory\_optimization} framework \cite{tum}. This method computes a racing line that minimizes integrated squared curvature, producing a smooth and dynamically feasible path within the given track boundaries.

\subsubsection{Track Representation}
The track is represented by its centerline $(x_c(s), y_c(s))$ and left/right boundaries $(x_l(s), y_l(s))$ and $(x_r(s), y_r(s))$, parameterized by arc length $s \in [0, S]$, where $S$ is the total track length. For optimization, the problem is formulated in the Frenet frame, where each point on the path is defined by:
\begin{equation}
    \mathbf{p}(s) = \mathbf{c}(s) + d(s) \cdot \mathbf{n}(s),
\end{equation}
where $\mathbf{c}(s)$ is the centerline position, $\mathbf{n}(s)$ is the unit normal vector to the centerline, and $d(s)$ is the lateral offset to be optimized.

\subsubsection{Curvature Cost Function}
The path curvature $\kappa(s)$ for a parametric path $\mathbf{p}(s) = (x(s), y(s))$ is computed as:
\begin{equation}
    \kappa(s) = \frac{\dot{x}(s)\ddot{y}(s) - \dot{y}(s)\ddot{x}(s)}{ \left[ \dot{x}(s)^2 + \dot{y}(s)^2 \right]^{3/2} },
\end{equation}
where $\dot{(\cdot)}$ and $\ddot{(\cdot)}$ denote first and second derivatives with respect to $s$.  
The optimization objective is to minimize the integrated squared curvature:
\begin{equation}
    J_{\kappa} = \int_{0}^{S} \kappa(s)^2 \, ds.
\end{equation}
This promotes smoothness, reduces lateral acceleration demands, and indirectly maximizes the achievable speed in curves.

\subsubsection{Optimization Problem}
The minimum curvature optimization problem is formulated as:
\begin{align}
    \min_{d(s)} \quad & \int_{0}^{S} \kappa(s)^2 \, ds \\
    \text{s.t.} \quad & d_{\text{min}}(s) \leq d(s) \leq d_{\text{max}}(s), \label{eq:trackbounds}\\
    & d(0) = d(S), \quad \dot{d}(0) = \dot{d}(S), \label{eq:lapclosure}
\end{align}
where \eqref{eq:trackbounds} enforces the track boundary constraints and \eqref{eq:lapclosure} ensures lap closure and continuity.

\subsubsection{Velocity Profile Generation}
Once the optimal path is obtained, a velocity profile is computed using the friction-limited acceleration constraint:
\begin{equation}
    a_x(s)^2 + a_y(s)^2 \leq (\mu g)^2,
\end{equation}
where:
\begin{equation}
    a_y(s) = v(s)^2 \kappa(s), \quad a_x(s) = \frac{dv(s)}{dt}.
\end{equation}
This yields the maximum feasible speed at each point:
\begin{equation}
    v_{\text{max}}(s) = \sqrt{\frac{\mu g}{|\kappa(s)| + \epsilon}},
\end{equation}
where $\epsilon$ is a small positive constant to avoid division by zero on straights.

The final reference trajectory $\mathcal{T}$ is thus:
\begin{equation}
    \mathcal{T} = \left\{ x(s), y(s), \kappa(s), v_{\text{max}}(s) \right\}_{s=0}^{S},
\end{equation}
which is exported as waypoints for the downstream Pure Pursuit controller.

\subsubsection{Integration with the Control System}
This precomputed minimum-curvature trajectory serves as a global reference, while our RL-augmented Pure Pursuit controller tracks it in real time. This combination leverages the global smoothness of the optimized path and the local adaptability of the RL-tuned lookahead distance.

\subsection{System Architecture}
Our system integrates a reinforcement learning (RL) agent with a geometric path tracking controller to achieve real-time adaptation in autonomous racing. The control pipeline consists of two core modules:

\begin{figure*}[t]
  \centering
  \includegraphics[
    width=.82\textwidth,     
    height=.70\textheight,   
    keepaspectratio,
    trim=12 12 12 12,clip    
  ]{\detokenize{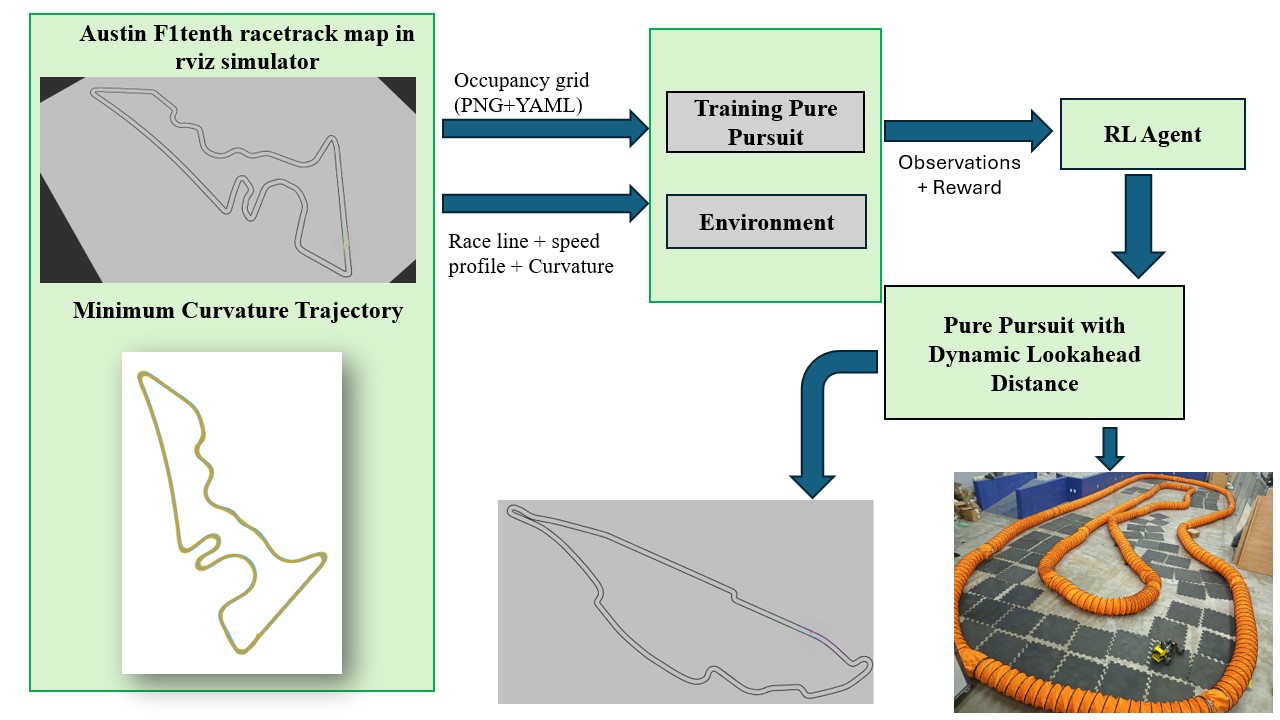}}
  \caption{Training-and-deployment workflow. A map and minimum-curvature raceline are generated offline and used by the simulator environment and baseline Pure Pursuit controller during training. The PPO agent receives observations and reward from the environment and outputs a dynamic lookahead policy, which is then integrated with Pure Pursuit for simulation validation and real-world deployment on a 1:10-scale autonomous racing platform.}
  \label{fig:overview}
\end{figure*}

\begin{enumerate}
    \item \textbf{RL-based Lookahead Distance Selection:} A PPO agent interacts with a Gym-wrapped Roboracer simulation environment, observing vehicle states and outputting an optimal lookahead distance $L_d$.
    \item \textbf{Pure Pursuit Control:} The selected $L_d$ is used to determine the target waypoint, from which the curvature and steering angle are computed.
    
\end{enumerate}

The vehicle operates in the RoboRacer simulator wrapped in a Gym-compatible API, enabling closed-loop training with synchronous ROS2 message exchange.
\Cref{fig:overview} shows the system overview. Starting from the Austin track map from the \texttt{roboracer\_racetracks} repository \cite{f1tenth_racetracks}, we visualize the occupancy grid in RViz and compute a minimum-curvature raceline using the TUM \texttt{global\_racetrajectory\_optimization} toolbox \cite{tum}. The resulting raceline provides curvature features and a speed profile capped at $12\,\text{m/s}$, which, together with the map, feed the training environment (simulator with sensing/odometry) and a baseline Pure Pursuit controller used during training. Both modules generate trajectories and observations for an RL agent (PPO), which learns a policy that selects a \emph{dynamic lookahead distance} online. 

The learned policy is integrated with Pure Pursuit in the Pure Pursuit with Dynamic Lookahead Distance module. We validate the controller in simulation and then deploy it on the real RoboRacer car. For completeness, the agent observes speed and multi-horizon curvature features derived from the raceline and outputs a continuous lookahead distance that is lightly smoothed before use; the detailed observation/action and reward definitions follow.

\subsection{Observation and Action Spaces}
The PPO agent observes a five-dimensional state vector at each timestep:
\begin{equation}
   s_t = \begin{bmatrix} v_t \\ \kappa_{0,t} \\ \kappa_{1,t} \\ \kappa_{2,t} \\ \Delta \kappa_t \end{bmatrix}
\end{equation}
where:
\begin{itemize}
    \item $v_t$: Vehicle speed in m/s, computed from odometry.
    \item $\kappa_{0,t}, \kappa_{1,t}, \kappa_{2,t}$: Absolute path curvatures at the closest, medium-horizon, and far-horizon waypoints, respectively.
    \item $\Delta \kappa_t = \kappa_{1,t} - \kappa_{0,t}$: Curvature change ahead of the vehicle.
\end{itemize}

The action space is continuous and one-dimensional:
\begin{equation}
    a_t = L_{t+1} \in [0.35, 4.0] \ \text{(meters)}
\end{equation}
representing the lookahead distance used by the Pure Pursuit controller. To reduce jitter, an exponential smoothing filter is applied before publishing $L_{t+1}$.
These bounds were chosen to cover the practical range of stable lookahead distances observed across low-speed cornering and high-speed straight-line driving in our platform and simulator setup.

\subsection{Reward Function Design}
The reward at timestep $t$ balances speed, smoothness, curvature minimization, collision avoidance, and path progress:
\begin{equation}
\begin{aligned}
R_t &= w_v v_t - w_e \left|L_t - L^*_t\right| - w_j \left|L_t - L_{t-1}\right| \\
&\quad - w_k \max(\kappa_{0,t},\kappa_{1,t},\kappa_{2,t}) \\
&\quad - w_c \cdot \mathbb{I}_{\text{collision}} + w_p \cdot \Delta p_t - w_s \mathbb{I}_{\text{stall}}
\end{aligned}
\end{equation}
where:
\begin{itemize}
    \item $L^*_t = 0.50 + 0.28 v_t - 3.5 \cdot \max(\kappa_{0,t},\kappa_{1,t},\kappa_{2,t})$: heuristic ideal lookahead.
    \item $\Delta p_t$: Waypoints advanced since the last step.
    \item $\mathbb{I}_{\text{collision}}$: True if the minimum LiDAR range $< 0.2$ m.
    \item $\mathbb{I}_{\text{stall}}$: True if $v_t < 0.05$ m/s for a duration.
\end{itemize}

The reward is clipped to:
\begin{equation}
    R_t \leftarrow \text{clip}(R_t, -20, 50),
\end{equation}
stabilizing training by limiting extreme outcomes.
The heuristic ideal-lookahead term is used only as a shaping signal to stabilize training; the deployed controller still relies on the learned policy output rather than a hand-crafted online rule.

\subsection{Pure Pursuit Control}
The Pure Pursuit algorithm selects a target point $(x_{\text{target}}, y_{\text{target}})$ that lies $L_{d,t}$ meters ahead along the path. Transforming to the vehicle frame yields coordinates $(x'_t, y'_t)$. The instantaneous curvature command is
\begin{equation}
    \gamma_t = \frac{2 y'_t}{L_{d,t}^2}.
\end{equation}
To avoid sudden steering changes, we apply a first-order low-pass filter:
\begin{equation}
    \bar{\gamma}_t = (1-\beta)\bar{\gamma}_{t-1} + \beta \gamma_t, \qquad \beta=0.4.
\end{equation}
The steering angle is then
\begin{equation}
    \delta_t = \arctan\!\big(L_w\, g(v_t)\, \bar{\gamma}_t\big),
\end{equation}
where $L_w$ is the wheelbase and $g(v_t)$ is the speed-dependent steering gain.

\subsection{Curvature-Aware Steering Gain}
The steering gain is dynamically adjusted based on vehicle speed:
\begin{equation}
    g(v) = \max\left(\min(m v + b, g_{\text{max}}), g_{\text{min}}\right)
\end{equation}
with:
\begin{equation}
    m = \frac{g_{\text{min}} - g_{\text{max}}}{v_{\text{max}} - v_{\text{min}}}, \quad
    b = g_{\text{max}} - m v_{\text{min}}
\end{equation}
Here, $g_{\min}$ and $g_{\max}$ denote the minimum and maximum steering-gain limits used to bound the speed-dependent schedule.
This prevents oversteering at high speeds and improves corner stability.

\subsection{Vehicle Dynamics Model}
The system assumes a kinematic bicycle model for state evolution:
\begin{align}
    \dot{x} &= v \cos(\theta) \\
    \dot{y} &= v \sin(\theta) \\
    \dot{\theta} &= \frac{v}{L_w} \tan(\delta) \\
    \dot{v} &= a
\end{align}
where $(x, y)$ is position, $\theta$ is heading, $v$ is speed, $a$ is acceleration, $\delta$ is steering angle, and $L_w$ is the wheelbase. This simplified model served as a practical approximation for the training and evaluation setup used in this work, while real-world validation was used to assess transfer beyond the model assumptions.

\subsection{Proximal Policy Optimization (PPO)}
We use the Stable-Baselines3 Proximal Policy Optimization (PPO) algorithm to train the policy $\pi_\theta(a_t | s_t)$ for adaptive lookahead tuning. PPO is a policy-gradient method that seeks to maximize a clipped surrogate objective, thereby improving training stability by preventing excessively large policy updates. Training uses:
\begin{itemize}
    \item Trajectory length $n_{\text{steps}} = 10{,}000$,
    \item Batch size $256$, 
    \item $n_{\text{epochs}}=5$, 
    \item Discount factor $\gamma = 0.99$, 
    \item GAE parameter $\lambda = 0.98$, 
    \item Clipping parameter $\epsilon = 0.2$, 
    \item target KL $=0.015$, 
    \item Entropy coefficient $0.02$ for exploration, 
    \item Value function coefficient $0.6$, and 
    \item Learning-rate schedule: a linear decay implemented as \texttt{learning\_rate(f)} $=\ell_0 f$ with $\ell_0=2.3927\times 10^{-4}$ (SB3 passes $f\!\in\![1,0]$ as remaining training progress).\footnote{Code: \texttt{initial\_lr = 0.00023927; learning\_rate = lambda f: initial\_lr * f}}
\end{itemize}

PPO maximizes the clipped surrogate:
\begin{IEEEeqnarray}{l}
\mathcal{L}_{\text{clip}}(\theta) = \nonumber\\[2pt]
\quad \mathbb{E}_t\!\Big[\min\!\big(r_t(\theta)\,\hat A_t,\,
\operatorname{clip}(r_t(\theta),1-\epsilon,1+\epsilon)\,\hat A_t\big)\Big]
\IEEEeqnarraynumspace\label{eq:ppo-clip}
\end{IEEEeqnarray}
with $r_t(\theta)=\pi_\theta(a_t\mid s_t)/\pi_{\theta_{\text{old}}}(a_t\mid s_t)$. The total loss combines policy, value, and entropy terms:
\begin{IEEEeqnarray}{l}
\mathcal{L}_{\text{ppo}}(\theta,\phi) = \nonumber\\[2pt]
\quad -\mathcal{L}_{\text{clip}}(\theta)
+ c_v\,\mathbb{E}_t\!\big[(V_\phi(s_t)-\hat R_t)^2\big]
- c_s\,\mathbb{E}_t\!\big[\mathcal{H}(\pi_\theta(\cdot\mid s_t))\big]
\IEEEeqnarraynumspace\label{eq:ppo-final}
\end{IEEEeqnarray}

Here, $\hat A_t$ denotes the generalized-advantage estimate, $\hat R_t$ the return target for value learning, $c_v$ the value-loss weight, and $c_s$ the entropy weight.

\subsection{Implementation and Training}
The training pipeline is implemented using \texttt{rclpy} in ROS2 and utilizes Stable-Baselines3’s PPO algorithm. Key training details include:
\begin{itemize}
    \item 800,000 environment steps
    \item VecNormalize for online normalization of observations and rewards
    \item TensorBoard logging for reward monitoring
    \item Evaluation every 10,000 steps with automatic best-model saving
\end{itemize}

\begin{figure}[t]
\centering
\begin{minipage}[t]{0.49\linewidth}\centering
  \includegraphics[width=\linewidth]{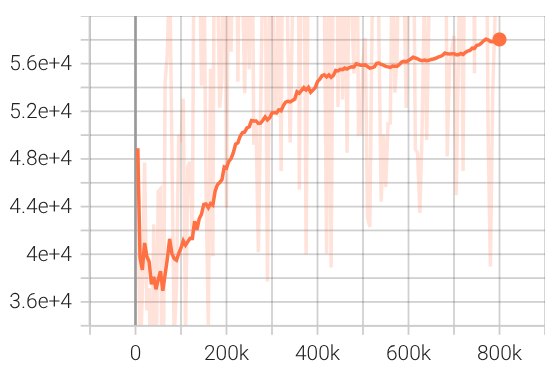}\\[2pt]
  {\footnotesize\bfseries Eval mean reward}
\end{minipage}\hfill
\begin{minipage}[t]{0.49\linewidth}\centering
  \includegraphics[width=\linewidth]{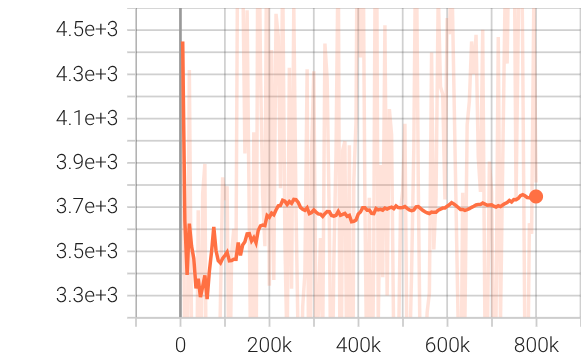}\\[2pt]
  {\footnotesize\bfseries Eval episode length}
\end{minipage}\\[4pt]
\begin{minipage}[t]{0.49\linewidth}\centering
  \includegraphics[width=\linewidth]{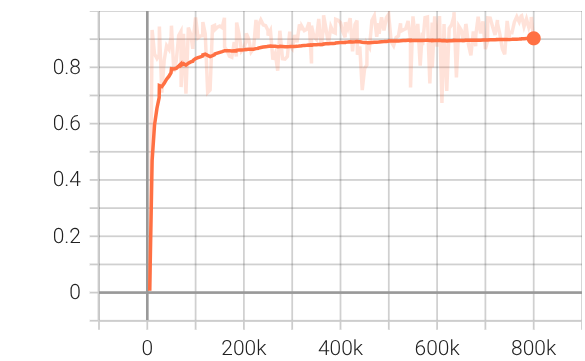}\\[2pt]
  {\footnotesize\bfseries Explained variance (critic)}
\end{minipage}\hfill
\begin{minipage}[t]{0.49\linewidth}\centering
  \includegraphics[width=\linewidth]{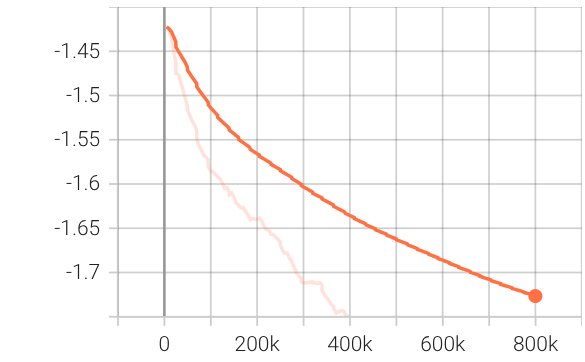}\\[2pt]
  {\footnotesize\bfseries Entropy}
\end{minipage}
\caption{Training diagnostics for PPO. Reward and episode length are computed on a fixed evaluation environment; explained variance and entropy are logged during training.}
\label{fig:ppo_tb}
\end{figure}


As shown in \Cref{fig:ppo_tb}, evaluation reward increases throughout training, accompanied by longer evaluation episodes and shrinking variance—indicating both performance and stability gains. The critic’s explained variance rises to $\sim$0.9, suggesting accurate value estimates that stabilize PPO updates. Meanwhile, entropy decreases smoothly, reflecting a controlled shift from exploration to exploitation consistent with our linear learning-rate decay. Together, these curves evidence a convergent policy that generalizes to the evaluation setting and yields robust, longer-horizon driving.

Training is conducted entirely in simulation using the F1TENTH Gym environment, with multiple runs to assess consistency and generalization. At each step, the PPO agent outputs $L_d$, which is smoothed and published to ROS. The Pure Pursuit node computes steering and speed commands, and the environment returns the next observation and reward. This closed-loop cycle enables the agent to learn curvature-aware, speed-adaptive lookahead selection for high-performance autonomous racing. All runs were executed for approximately 4~hours on a Dell Precision~3660 workstation (Intel Core i9-13900, 32~GiB RAM), running Ubuntu~22.04.5~LTS (64-bit, X11); graphics were reported as “NVIDIA Corporation / Mesa Intel Graphics (RPL-S)”.

\section{Experimental Results}
\label{sec:exp}

We evaluate the proposed method in simulation and present initial real-car validation on a RoboRacer vehicle.

\par\addvspace{0.7\baselineskip}

\paragraph*{Fixed-Lookahead Failure Modes}
A Pure Pursuit (PP) controller with a \emph{fixed} lookahead $L$ is highly map- and speed-profile dependent. When $L$ is too small, the vehicle exhibits oscillations on straights and may cut inside on curves; when $L$ is too large, tracking is steadier on straights but it understeers and fails in tighter corners. In practice, each new map demands extensive retuning of $L$ (and of any hand-crafted ``dynamic'' heuristics that couple $L$ to speed), and success is not guaranteed.

\par\addvspace{0.7\baselineskip}

\paragraph*{Setup and zero-shot evaluation}
We train the Dynamic-$L$ policy with PPO entirely in simulation on the Austin racetrack and evaluate it without finetuning on two unseen tracks, Montreal and Yas Marina, from the \texttt{roboracer\_racetracks} repository \cite{f1tenth_racetracks}. We compare three controllers: fixed-lookahead Pure Pursuit, a hand-crafted adaptive Pure Pursuit baseline, and the proposed RL-based Dynamic-$L$ controller. To increase evaluation difficulty, we tested the controllers under accelerated raceline speed profiles on the unseen tracks whenever stable repeated lap completion could be achieved.

Because each controller tolerated different maximum speed-profile scalings on the unseen tracks, the reported results should be interpreted primarily as evidence of robustness and reduced retuning burden rather than as a strict like-for-like comparison under identical operating conditions.

\paragraph*{Adaptive Pure Pursuit baseline}
For the adaptive Pure Pursuit baseline, the lookahead distance is scheduled online as
\[
L_d = \mathrm{clip}(a + b v_t,\, L_{\min},\, L_{\max}),
\]
where \(v_t\) is the vehicle speed. We use the same feasible lookahead bounds as the learned controller, namely \(L_d \in [0.35, 4.0]\) m. The same scheduling rule and parameter values are used across all simulator test tracks, without additional per-track retuning. This baseline is included to distinguish the benefit of learned context-dependent adaptation from the benefit of dynamic lookahead scheduling alone. The adaptive baseline is evaluated in simulation only and is not included in the present hardware comparison.

\paragraph*{Interpretability and robustness evidence}
We use robustness to mean the ability to complete laps reliably without per-map retuning when track geometry and speed profiles change, and when transferring from simulation to hardware. Accordingly, we evaluate (i) zero-shot generalization from the training map to two unseen maps under accelerated speed profiles, and (ii) deployment on a real RoboRacer vehicle under an independently generated raceline speed profile. The inclusion of an adaptive heuristic Pure Pursuit baseline further tests whether the gains of the proposed method arise from learned context-dependent adaptation rather than from dynamic lookahead scheduling alone.

\par\addvspace{0.7\baselineskip}

\paragraph*{Montreal RoboRacer racetrack quantitative results (10 consecutive laps)}
On Montreal, Dynamic-$L$ completes 10/10 laps under a \textbf{+13\%} speed profile, achieving a mean lap time of \textbf{33.16\,s} with a standard deviation of \textbf{0.069\,s}, as reported in Table~\ref{tab:montreal-times}. The adaptive Pure Pursuit baseline also completes 10/10 laps, but under a slower \textbf{+10\%} speed profile, recording \textbf{34.15\,s} with a standard deviation of \textbf{0.061\,s}. The fixed-\(L\) PP baseline (with \(L=1.2\,\text{m}\)) completes 10/10 laps only after reducing the speed profile by \textbf{-10\%}, where it records \textbf{41.36\,s} with a standard deviation of \textbf{0.067\,s}. Lap-time statistics are reported in Table~\ref{tab:montreal-times}.

These results indicate that the learned dynamic-lookahead policy reduces sensitivity to per-map tuning and supports reliable operation at a more aggressive speed profile than either baseline on this unseen track.

\begin{table}[t]
\caption{Montreal track lap times over 10 consecutive laps (s).}
\centering
\setlength{\tabcolsep}{6pt}
\begin{tabular}{lcccc}
\toprule
Controller & Mean & Std & Min & Max \\
\midrule
\(RL\) PP (\(+13\%\) speed) & \textbf{33.16} & 0.069 & 33.06 & 33.28 \\
Adaptive PP (\(+10\%\) speed) & 34.15 & 0.061 & 34.02 & 34.23 \\
Fixed-\(L\) PP (\(-10\%\) speed) & 41.36 & 0.067 & 41.25 & 41.45 \\
\bottomrule
\end{tabular}
\label{tab:montreal-times}
\end{table}

 \begin{table}[t]
\caption{Yas Marina track lap times over 10 consecutive laps (s).}
\centering
\setlength{\tabcolsep}{6pt}
\begin{tabular}{lcccc}
\toprule
Controller & Mean & Std & Min & Max \\
\midrule
\(RL\) PP (\(+15\%\) speed) & \textbf{46.05} & 0.035 & 46.00 & 46.10 \\
Adaptive PP (\(+14\%\) speed) & 46.79 & 0.095 & 46.67 & 46.79 \\
Fixed-\(L\) PP (orig.\ speed) & 52.81 & 0.18 & 52.52 & 53.10 \\
\bottomrule
\end{tabular}
\label{tab:yas-times}
\end{table}

\begin{table}[t]
\caption{Real-car lap times over 10 consecutive laps with a raceline speed profile capped at \(5~\text{m/s}\) (s).}
\centering
\setlength{\tabcolsep}{6pt}
\begin{tabular}{lcccc}
\toprule
Controller & Mean & Std & Min & Max \\
\midrule
\(RL\) PP  & \textbf{11.13} & 0.26 & 10.79 & 11.66 \\
Fixed-\(L\) PP  & \multicolumn{4}{c}{DNF / failed to complete repeated laps} \\
\bottomrule
\end{tabular}
\label{tab:realcar-times}
\end{table}

\begin{figure}[t]
  \centering
  \makebox[\linewidth][c]{%
    \includegraphics[width=1.04\linewidth]{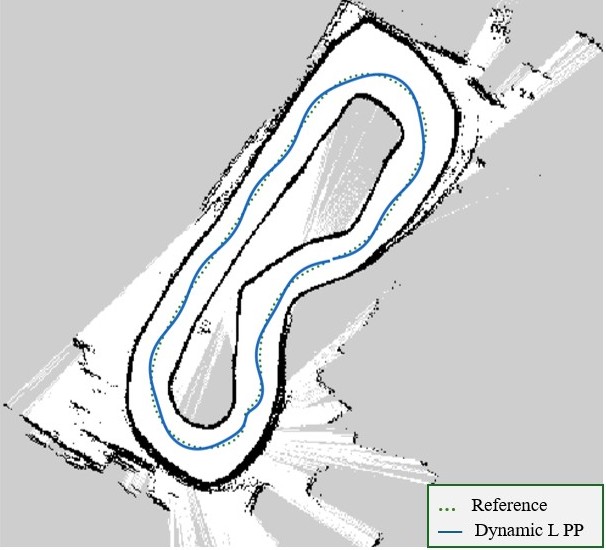}}
  \caption{Qualitative real-car trajectory overlay. The learned Dynamic-$L$ controller tracks the reference raceline and completes repeated laps on the mapped track under the tested speed profile.}
  \label{fig:real-overlays}
\end{figure}

\par\addvspace{0.7\baselineskip}

\paragraph*{Yas Marina RoboRacer racetrack quantitative results (10 consecutive laps)}
On Yas Marina, Dynamic-$L$ completes 10/10 laps under a \textbf{+15\%} accelerated speed profile, achieving a mean lap time of \textbf{46.05\,s} with a standard deviation of \textbf{0.035\,s}, as reported in Table~\ref{tab:yas-times}. The adaptive Pure Pursuit baseline also completes 10/10 laps, but under a slightly lower \textbf{+14\%} speed profile, recording \textbf{46.79\,s} with a standard deviation of \textbf{0.095\,s}. For the fixed-\(L\) baseline (with \(L=1.35\,\text{m}\)), stable repeated completion at the accelerated profile was not achieved; therefore, we report its results under the original speed profile, where it records \textbf{52.81\,s} with a standard deviation of \textbf{0.18\,s}. Lap-time statistics are reported in Table~\ref{tab:yas-times}.

This comparison further supports that the learned dynamic-lookahead policy reduces retuning burden and tolerates a more aggressive speed profile than the fixed-lookahead baseline while also achieving a lower lap time than the adaptive scheduling baseline under a slightly more aggressive speed profile on this unseen track.



\par\addvspace{0.7\baselineskip}



\par\addvspace{0.7\baselineskip}
\paragraph*{Real-car validation}
We additionally validated the proposed Dynamic-$L$ controller on a real 1:10-scale autonomous racing vehicle using a raceline speed profile capped at \(v_{\max}=5~\text{m/s}\) on a mapped track. On hardware, Dynamic-$L$ completed 10/10 laps with a mean lap time of \(11.13 \pm 0.26\) s (range \(10.79\)--\(11.66\) s). Under the same profile, a fixed-\(L\) Pure Pursuit controller failed to complete repeated laps reliably. Fig.~\ref{fig:real-overlays} qualitatively shows that Dynamic-$L$ tracks the reference raceline and completes full laps under the tested profile. We limit the hardware comparison to the fixed-lookahead baseline, while the adaptive Pure Pursuit baseline is evaluated in simulation only.

\section{Conclusion}
We introduced a PPO-based \emph{dynamic lookahead} policy $\pi_\theta(L_d\mid s)$ for Pure Pursuit. Trained in simulation on one track and evaluated zero-shot on unseen tracks, the policy completed repeated laps reliably under aggressive speed profiles while preserving the simplicity and interpretability of the underlying geometric controller. Compared with fixed-lookahead Pure Pursuit and a hand-crafted adaptive Pure Pursuit baseline, the proposed method demonstrated stronger robustness to track changes and reduced retuning burden. On Montreal and Yas Marina, Dynamic-$L$ achieved \(33.16 \pm 0.069\)\,s and \(46.05 \pm 0.035\)\,s, respectively, while operating under \(\,+13\%\) and \(\,+15\%\) speed profiles. In both cases, the learned policy achieved the best lap times among the tested settings while also tolerating equal or more aggressive speed-profile scaling than the baselines. These results suggest improved robustness to track changes and reduced retuning burden, rather than a strict like-for-like comparison under identical operating conditions.

\emph{Real-car validation.} Experiments further support sim-to-real transfer of the learned dynamic-lookahead policy on a 1:10-scale autonomous racing platform. Future work will expand the hardware study to include broader adaptive scheduling rules and additional learning-based racing controllers and will further analyze which components of the present design—state features, smoothing, steering-gain scheduling, and reward terms—are most critical to performance.

\bibliographystyle{IEEEtran}
\bibliography{refs}
\end{document}